# A Multi-Objective Deep Reinforcement Learning Framework

Thanh Thi Nguyen[1], Ngoc Duy Nguyen[2], Peter Vamplew[3], Saeid Nahavandi[2], Richard Dazeley[1], Chee Peng Lim[2]

[1]School of Information Technology, Deakin University, Victoria, Australia
[2]Institute for Intelligent Systems Research and Innovation, Deakin University, Victoria, Australia
[3]School of Science, Engineering and Information Technology, Federation University, Australia

E-mail: thanh.nguyen@deakin.edu.au; Tel: +61 3 52278281.

**Abstract**
This paper introduces a new scalable multi-objective deep reinforcement learning (MODRL) framework based on deep Q-networks. We develop a high-performance MODRL framework that supports both single-policy and multi-policy strategies, as well as both linear and non-linear approaches to action selection. The experimental results on two benchmark problems (two-objective deep sea treasure environment and three-objective Mountain Car problem) indicate that the proposed framework is able to find the Pareto-optimal solutions effectively. The proposed framework is generic and highly modularized, which allows the integration of different deep reinforcement learning algorithms in different complex problem domains. This therefore overcomes many disadvantages involved with standard multi-objective reinforcement learning methods in the current literature. The proposed framework acts as a testbed platform that accelerates the development of MODRL for solving increasingly complicated multi-objective problems.

## 1. Introduction

Most multi-objective reinforcement learning (MORL) studies so far have been on relatively simple gridworld tasks, so extending current algorithms to solve more complex problem domains is important. The current algorithms such as tabular Q-learning (Watkins and Dayan, 1992) require a high degree of memory usage, which is inefficient and impractical when the environment's state space is large. Deep reinforcement learning (DRL) approaches are possible solutions to overcome this problem because the memory is only required to store the neural network or experience replay.

MORL methods have been extensively applied to solve real-world problems. For example, Zhou et al. (2019) proposed a molecule deep Q-network (DQN) and extended it to multi-objective scenarios where drug-likeness is maximized while similarity between molecules needs to be maintained at the same time. Li and Czarnecki (2019) applied a multi-objective DQN approach to autonomous driving where multiple aspects are required to consider concurrently such as obeying traffic rules, avoiding collision, reaching destination as fast as possible, and also ensuring the safety of passengers. Likewise, Hasanvand et al. (2020) considered a multi-objective environmental problem where zero-emission and cost-effective energy scheduling are both achieved simultaneously.

There has been a small amount of prior work investigating deep methods for MORL, henceforth multi-objective deep reinforcement learning (MODRL) problems. As such, no standard benchmarks have yet emerged. Vamplew et al. (2017) developed an MORL framework based on RL_Glue (Tanner and White, 2009), known as MORL_Glue, in which they have implemented benchmark environments and several tabular and tile-coding MORL algorithms. However, this framework currently only provides support for passing state information in the form of vectors of integer or continuous values but does not allow passing images or other complex state representations. In addition, the implementation of the environments in MORL_Glue does not generate image-based representations of the state. More importantly, this framework does not support implementation of deep learning algorithms, e.g. DQN (Mnih et al., 2015) or its variants.

Traditional MORL methods normally cannot take image inputs because the encoded data are too large. Therefore, the integration of deep methods into the MORL problems is important. It is desirable to design a modularized and scalable MODRL framework to deal with increasingly complicated problems. To this aim, we propose a benchmark Python framework that supports both single-policy and multi-policy approaches to solving MODRL problems. Our framework is generic and highly modularized so that it can accommodate any DRL method. In this paper, we use DQN for demonstration. To enhance flexibility, we design the framework based on a combination of the Tensorflow library (Abadi et al., 2016) and a concept of network configuration. Using the network configuration, the framework flexibly accepts any state representation (scalar, vector, and graphical data) and simplifies the modification of different DRL algorithms by applying a plug-and-play method. Furthermore, our framework uses multithreading to significantly reduce the training time; therefore, it is highly effective in multi-policy cases. Extensive experimental results on two multi-objective problems, i.e. Deep Sea Treasure and MO-Mountain Car, demonstrate the effectiveness of the framework.

In summary, the paper has the following contributions. It proposes one of the first frameworks to facilitate the development of MODRL algorithms. Our MODRL framework supports single and multi-policy strategies, and linear and non-linear approaches to action selection. In addition, the framework proposes the use of hypervolume computation for evaluating different MODRL algorithms and a set of graphical environments, which can be used as a testbed to compare MODRL methods. As opposed to existing works, our framework would serve as a fundamental unit to support future theoretical studies of MODRL methods as well as accommodate their implementations and applications to solve real-world problems.

Before presenting the development of the MODRL framework, we present an overview of MORL methods in the next section. Section 3 describes the single-policy and multi-policy approaches implemented in the proposed framework. Experiments and discussions are presented in Section 4, which is followed by conclusions and further work in Section 5.

## 2. MORL Methods and Deep Learning Extensions

### *2.1. Overview of MORL Methods*

Many decision making problems in the real world requires the consideration of more than one objective. MORL extends the conventional single-objective RL methods to characterize two or more objectives simultaneously. The reward signal of MORL is not a scalar but a vector where each element corresponds to an objective. If the objectives are independent or directly related, they can be combined into a single objective, and optimizing the resulting objective can solve the problem. However, the objectives of MORL are often conflicting so that maximization of one objective normally leads to minimization of another. This is a more challenging scenario where trade-offs among objectives need to be considered. Evaluation of MORL algorithms is therefore often based on a Pareto front, which represents compromise solutions among the objectives. For a detailed discussion on MORL, we refer the readers to the survey by Roijers et al. (2013).

Current MORL methods can be classified into two categories: single-policy (Van Moffaert et al., 2013; Hein et al., 2017) and multi-policy methods (Van Moffaert and Nowé, 2014; Pirotta et al., 2015; Parisi et al., 2016, Ruiz-Montiel et al., 2017). Single-policy methods attempt to find a single solution of the problem whilst multi-policy methods can find multiple solutions simultaneously. Single-policy methods have an advantage that requires less computational expenses as compared with that of multi-policy methods. They, however, require prior information about the objective preference from the user. This may lead to a solution that is undesired by the user because a small change of the objective preference may produce significant variations of the solution. Multi-policy methods can generate multiple solutions to approximate the true Pareto front so that users can select suitable solutions that satisfy their varied preferences. Presenting the optimal front to users provides them the trade-off information as well as the interaction among the competing objectives. The main disadvantage of generating multiple policies is the high computational cost, which impacts the online learning capability of MORL algorithms. Therefore, single policy methods are normally carried out in the context of online learning.

In MORL, the rewards learned by the agent are represented by a vector and the action is chosen by applying a greedy selection paradigm that takes into account these reward vectors corresponding to the actions. There are several algorithms proposed in the literature for solving MORL problems, most of which are based on the scalarisation method to transform the multi-objective problem to a single objective one (Vamplew et al., 2008). The scalarisation can be linear (Castelletti et al., 2013; Khamis and Gomaa, 2014; Ferreira et al., 2017) or nonlinear (Tesauro et al., 2008). Other methods include the advanced version of the two-phase local search (Van Moffaert et al., 2014), analytic hierarchy process, geometric, ranking, convex hull, and varying parameter approaches (Liu et al., 2015). Notably, Gábor et al. (1998) introduced a thresholded lexicographic ordering (TLO) that maximises the performance on one chosen objective subject to meeting a threshold level of performance on the remaining objectives. In addition, Perny and Weng (2010) defined a scalarisation function as the distance from a specified target point in the objective space (the Tchebycheff distance), whereas Van Moffaert et al. (2013) tackled scalarisation using the hypervolume metric.

In this paper, we implement two approaches: the linear weighted sum and the nonlinear TLO method for demonstration purposes. For some problem domains, linear methods may be inadequate to accurately or easily express the desired trade-offs among the objectives (Roijers et al., 2013), and non-linear methods may be preferable (Gábor et al., 1998; Issabekov and Vamplew, 2012). In our framework, for the problem domains where the linear method is unable to find the optimal policies, we propose the use of TLO as an alternative approach. The TLO method is implemented based on thresholding and lexicographic ordering of the objective values of the available actions. We apply thresholds to the first $N-1$ objectives and leave the last objective unconstrained. The thresholds can be estimated based on the observation of minimum and maximum Q-values of the objectives. Details of the TLO method are as follows.

Define $Q_{s,a,j}$ as the value of the $j$th objective of action $a$ at state $s$, and $T_j$ as the threshold of objective $j$th, the truncation yields:

$$TQ_{s,a,j} = \min(Q_{s,a,j}, T_j) \quad (1)$$

At a given state $s$, the algorithm selects the greedy action $a'$ if the following recursive function superior($TQ_{s,a',1}$, $TQ_{s,a,1}$, 1) returns True $\forall a \in A$ where $A$ is the set of available actions (Vamplew et al., 2011).

function superior($TQ_{s,a',i}$, $TQ_{s,a,i}$, $i$):

If $TQ_{s,a',i} > TQ_{s,a,i}$ Then

```
        Return True
    Else If TQ_{s,a',i} = TQ_{s,a',i} Then
        If i < N Then
            Return superior(TQ_{s,a',i+1}, TQ_{s,a,i+1}, i + 1)
        Else
            Return True
    Else
        Return False
```

*2.2. Current MODRL Methods*

Mossalam et al. (2016) extended the DQN (Mnih et al., 2015) to handle single-policy linear MORL. They addressed the multi-policy task of finding the convex coverage set (CCS - the complete set of policies such that an optimal policy is available for any possible weight vector) by embedding their DQN algorithm within an outer loop method, which identifies the weight vectors to use in training so as to establish the CCS. They used two small gridworld tasks in two different fashions as the test problems of MODRL. First, they provided the underlying discrete or continuous state information directly to the deep neural network (DNN) – this information is low-dimensional so the capacity of the DQN is essentially overkill for such tasks. The second approach is a better evaluation of MODRL methods, as they use a visualization of the environment to generate an image for input to the DNN. They showed that efficiencies can be achieved by retaining parts, but not all, of the DNN when the outer loop changes the weights. Overall this method is able to address the multi-policy linear MORL problem, but it is doing so via sequential rather than parallel learning of these policies.

Tajmajer (2017) also extended the DQN, but used a non-linear action selection approach based on a subsumption architecture. A prioritized ordering of objectives is specified, and higher priority objectives can 'suppress' the Q-values associated with lower-priority objectives. The suppression values are state-dependent so the whole system essentially performs a dynamic, state-dependent weighting of the Q-values whenever an action is selected. That work addresses the single-policy non-linear MORL problem, but in a manner which is tied to one specific form of non-linear action selection.

Recent notable study on MODRL was the introduction of diverse experience replay that overcomes the inherent non-stationary problem of an experience replay memory while using a dynamic weight setting, i.e., the relative importance weights between different objectives can be changed over time (Abels et al., 2018). Alternatively, Tajmajer (2018) proposed a modular architecture such that multiple DQNs control the agent's behaviour in real time. To make it feasible, decision values are used to scalarize different DQNs into a unique action. On the other hand, Wang et al. (2019) designed an end-to-end multi-objective workflow that enables scheduling over infrastructure-as-a-service clouds and supports multiple agents.

*2.3. MODRL Framework Development*

*2.3.1. Framework architecture*

We design the framework so that it can easily apply different DRL algorithms to a variety of problem domains. To fulfil this constraint, the design of the framework architecture must be modularized to separate the cohesion between three essential components: neural networks, DRL algorithms, and environments. Fig. 1 presents a high-level system architecture of the framework.

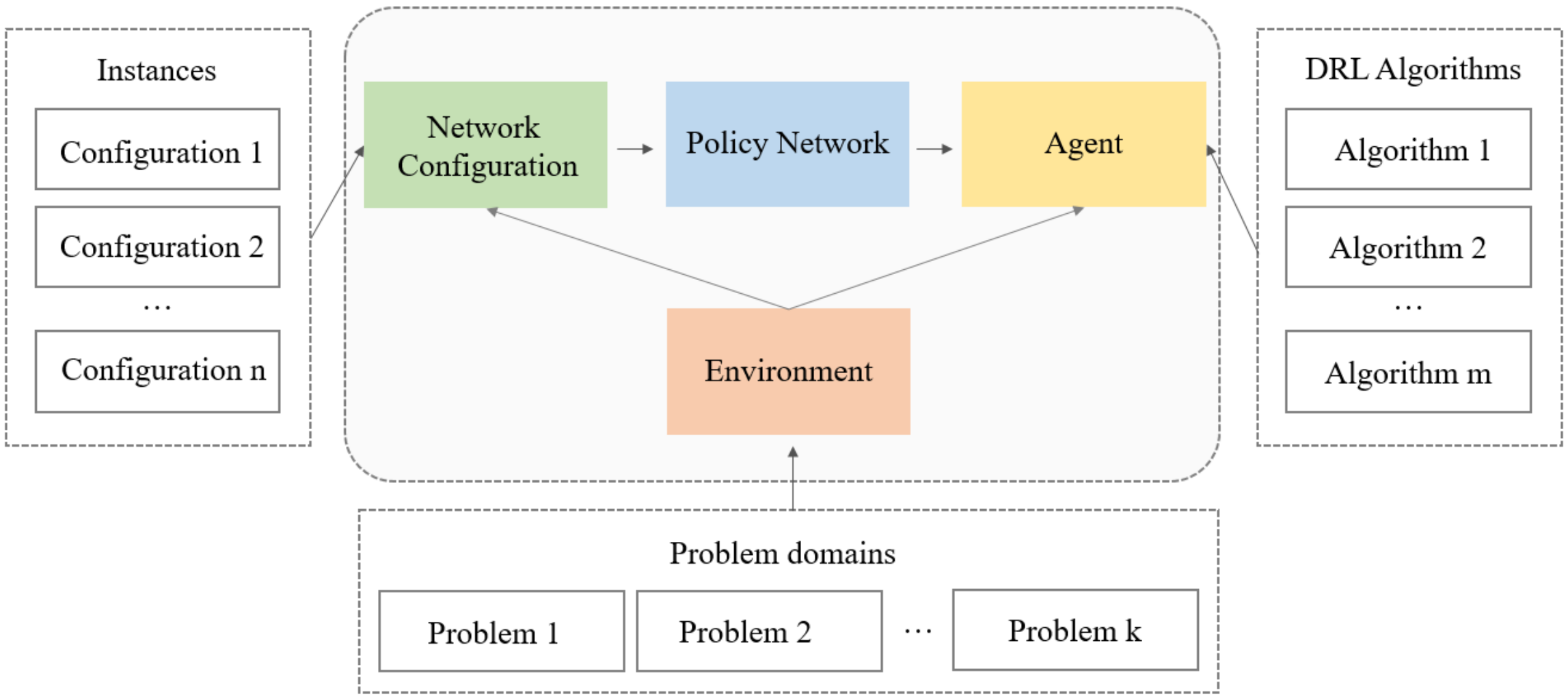

Fig. 1. A high-level system architecture of the MODRL framework.

As shown in Fig. 1, the network configuration component plays an important role to configure the neural networks. Examples of such configurations include number of layers, the existence of convolutional neural networks, linear or nonlinear weights, and state representation. By varying predefined parameters of network configuration, we can create an instance of network configuration that suits our needs. The configuration is then used by a policy network module to generate the required network (computing graph). Finally, a DRL algorithm (an agent) uses the neural network together with environment information to perform the training or evaluation task. Note that all defined problems must follow a common interface given by the environment component. To evaluate another DRL algorithm, we simply implement a different agent and use it in place of the existing one while other components are kept unchanged. Therefore, the proposed framework is highly modularized. We also utilize the multithreading technique to maximize performance throughput of the algorithms, especially in multi-policy tasks. In this paper, for demonstration, we implement two problems, namely Deep Sea Treasure and Mountain Car; the DRL algorithm is DQN; and two instances of network configuration are used: one for the linear method and another for the nonlinear TLO method. Details of the framework implementation can be referred to http://www.deakin.edu.au/~thanhthi/drl.htm.

*2.3.2. Single-policy DQN*

The linear approach is the most straightforward extension of DRL to MODRL. It involves learning a single policy based on a linear scalarisation of the objectives using a fixed set of weights. This is equivalent to learning the optimal policy for a single-objective Markov decision process for which the objectives are pre-scalarized into a single reward. In our framework based on the DQN, the agent receives a vector of rewards at each time step, not a scalar value. In addition, the agent is provided with a fixed weight vector, $w$, indicating the relative desirability of different objectives. As the weights of this scalarisation are fixed, they are not included as inputs to the DQN.

Given a weight vector $w = \{w_1, w_2, \ldots, w_n\}$ and reward vector $r = \{r_1, r_2, \ldots, r_n\}$, the loss function of multi-output DQN is defined as follows:

$$L(\theta) = \sum_{i=1}^{n} L_i(\theta) \quad (2)$$

in which

$$L_i(\theta) = E\left(\left(\gamma \max_{a'} Q_i(s', a'; \theta') - Q_i(s, a; \theta)\right)^2\right) \quad (3)$$

where $\gamma$ is the discounted rate, $0 \leq \gamma \leq 1$, and $s, s', a, a', \theta$, and $\theta'$ denote the current state, next state, current action, next action, estimation network's weights and target network's weights, respectively.

Fig. 2 depicts the network configuration used in our proposed framework, which includes 3 convolutional layers. The first convolutional layer is characterized by 32 filters of size 8x8 and a stride of 4. The second layer has 64 filters of size 4x4 and a stride of 2. The last layer has 64 filters of size 3x3 and a stride of 1. These convolutional layers are followed by a 512-unit fully connected layer. The ReLU function activation is used in all of the units. Finally, the output layer includes multiple groups of nodes where the number of groups is equal to the number of objectives. Each group consists of a number of nodes corresponding to the number of possible actions.

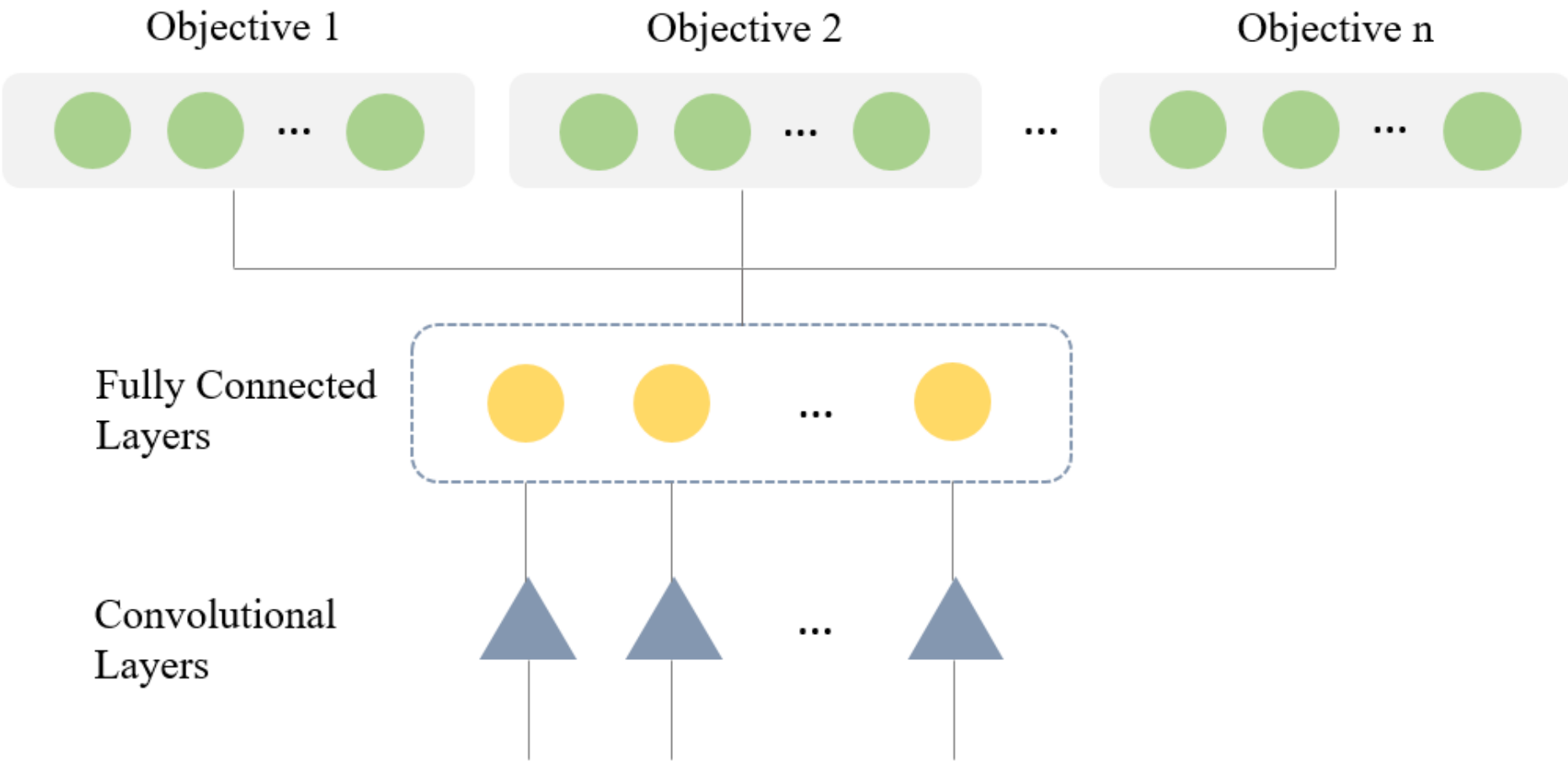

Fig. 2. Neural network structure used in the proposed DQN-based MODRL framework.

*2.3.3. Multi-policy DQN*

The choice of weights in a linear scalarisation is intended to represent the desirable trade-offs between different objectives. In many problems, the user's preferences pertaining to the objectives may change over time. The single-policy approach described in subsection 2.3.2 requires the agent to re-learn a new policy whenever the weights change, which can introduce unwarranted delays in responding to changes, particularly if the agent is operating in a real-time context. In our framework, we implement multiple threads to allow the agents to learn in parallel multiple policies, such that it has an optimal policy in advance for any possible set of weights (linear weighted sum) or thresholds (nonlinear method) which it might encounter. This reduces a significant amount of training time to achieve concurrent optimal policies for any possible sets of weights or thresholds. In this way, it can immediately adapt its behaviours when a change in weights or thresholds occurs.

## 3. Experiment Settings and Evaluations

There are several benchmark environments to test MORL algorithms such as Deep Sea Treasure (DST), MO Puddle World, MO Mountain Car, and Resource Gathering (Vamplew et al., 2011). In this paper, we evaluate our proposed MODRL framework using the DST and MO Mountain Car problems because they have different numbers of objectives, which ensure a general conclusion driven out from the experimental results. The DST environment has two objectives whilst the Mountain Car problem has three objectives. Each objective is characterised by a reward signal that can be either intrinsic or extrinsic (Uchibe and Doya, 2008). The intrinsic reward takes a non-zero signal most of the time, e.g. the time penalty for each time step. In contrast, the extrinsic reward only gets non-zero signal at specific time such as when the goal state is reached. The DQN parameter settings used in our experiments are presented in Table 1. These parameters are selected by using grid search so that the problem's convergence occurs in the minimum number of steps.

Several metrics are available to evaluate the performance of MORL algorithms such as hypervolume indicator, accumulated reward, regret metric, user-based testing or simulated user testing (Vamplew et al., 2011). The hypervolume metric is used to measure the performance of MODRL problems in this study because it can provide a single value to compare different learning algorithms and it does not require knowledge about the true optimal front or its approximation.

Table 1. DQN settings for our experiments

| **Parameters** | **Values** | | |
|---|---|---|---|
| Initial epsilon | 1.0 | | |
| Final epsilon | 0 | | |
| Learning rate | 0.0001 | | |
| Gamma (discounted rate) | 0.9 | | |
| Target network update | 1000 steps | | |
| Root mean square (RMS) optimizer | decay = 0.99, epsilon = 1e-6 | | |
| Benchmark environments | 3-column DST | 5-column DST | MO Mountain Car |
| Action repeat | 1 | 1 | 5 |
| Epsilon annealing steps | 96,000 | 190,000 | 200,000 |
| Experience replay size | 50,000 | 100,000 | 20,000 |
| Warmup steps | 5,000 | 10,000 | 2,000 |
| Training steps | 100,000 | 200,000 | 200,000 |

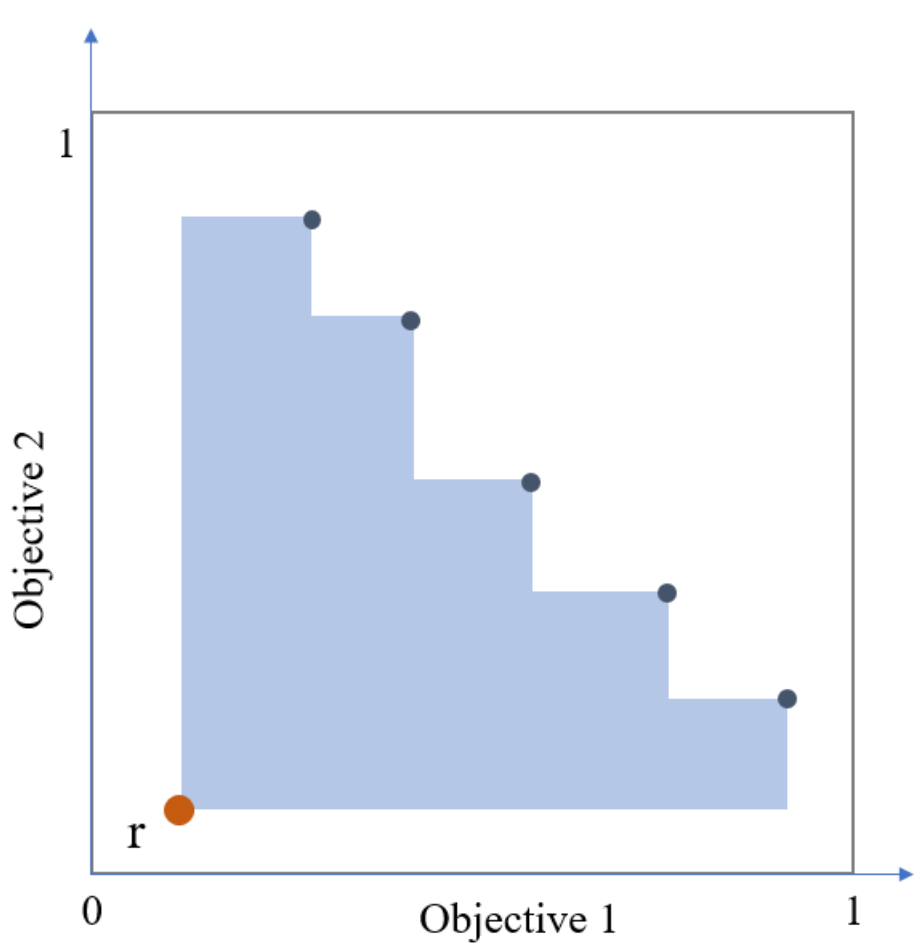

Fig. 3. The hypervolume is derived by the blue region, bounded by the optimally approximated front and the reference point, $r$.

Fig. 3 illustrates the hypervolume indicator in a two-objective environment with a reference point, $r$. The reference point must be chosen as to be dominated by all members of a frontal set $S$, and it must be the same when evaluating the performance of different learning algorithms. The larger the volume the better the algorithm is. The advantage of hypervolume is that any improvement in terms of accuracy, extent or diversity of the frontal set can be reflected by a larger hypervolume value.

To analyse and compare the performance of different algorithms, we measure the hypervolume of the approximated fronts not only after convergence but also during the learning process of the algorithms. To obtain the online hypervolume, the exploration is turned off and one run is made through the environment until the episode is finished. After that, the exploration is turned back on, and the algorithm’s learning process proceeds normally. For single-policy methods, we run multiple trials one after another with different weights or thresholds, logging the rewards received at fixed intervals during training and then merging the logged results to produce a set of values which can be used to calculate the hypervolume. For multi-policy methods, we run multiple threads to find multiple solutions in parallel; therefore, the hypervolume can be computed using these intermediate solutions.

**4. The Deep Sea Treasure (DST) Problem**

DST takes advantages of predefined Pareto solutions so that it becomes a normative multi-objective environment to verify new methods. While the original DST proposed in Vamplew et al. (2011) had a fixed spatial structure and rewards, here we use the generalised DST described in Vamplew et al. (2017) which allows the creation of parameterised instances, varying in terms of the size of the state space, and the values of the treasure rewards. The state output of DST can be a scalar (current position of the agent) or a graphical representation (image). Two grid DST environments are illustrated in Fig. 4, with the dimensions of 7x3 and 10x5, respectively.

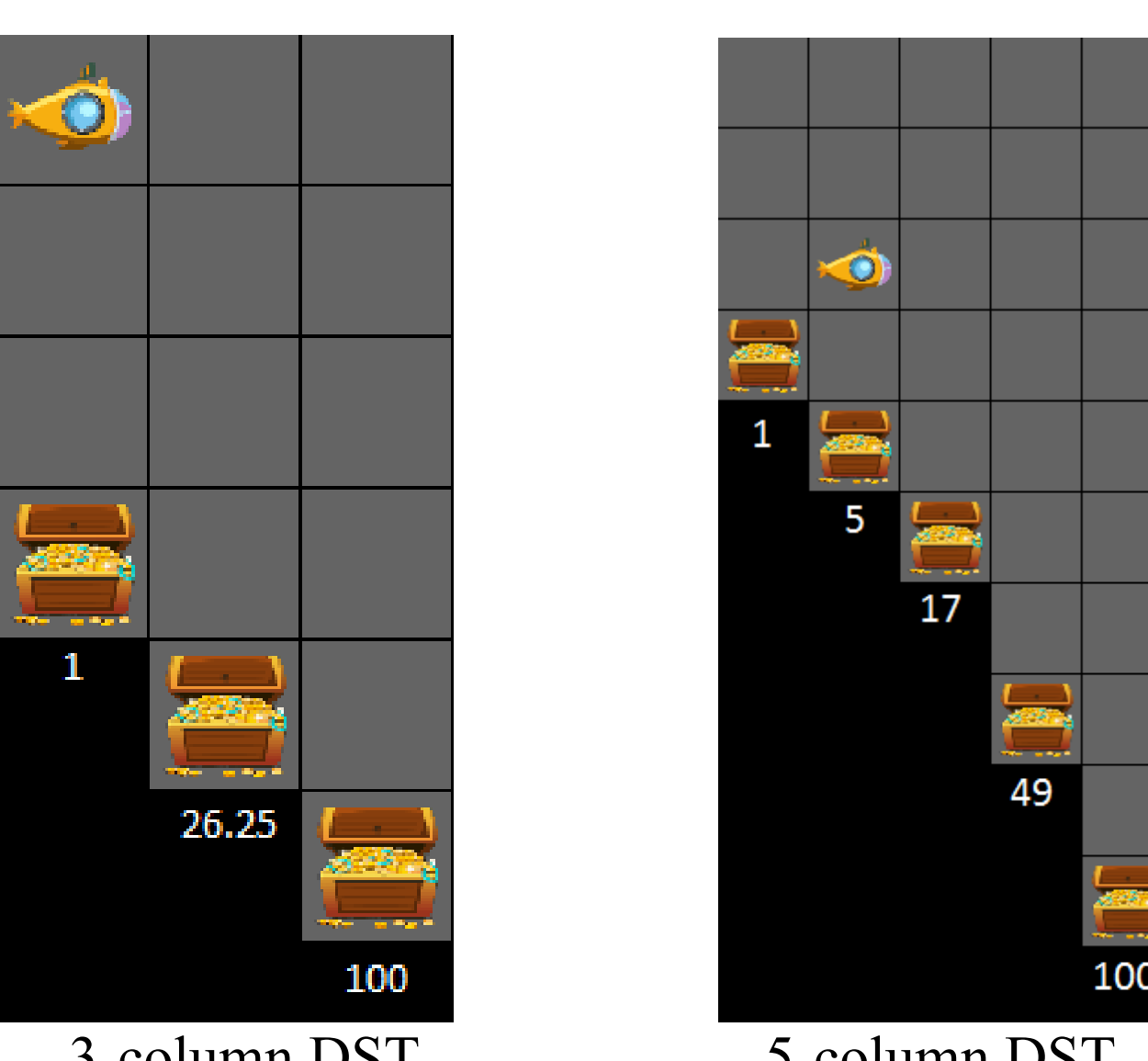

3-column DST 5-column DST

Fig. 4. Two experimental Deep Sea Treasure environments. The numbers below the treasures show their corresponding values.

The agent is designed to control a submarine that searches for treasure under a sea. Two objectives need to be optimized: maximize the treasure values and minimize the search time. Therefore, the DST problem has one extrinsic (treasure values) and one intrinsic (time penalty) reward. The submarine starts each episode at the top left state and ends when it finds a treasure location, or the predefined maximum number of actions is reached. Four actions including move up, down, left, and right are available to the agent. The agent receives a reward characterized by a 2-element vector representing the treasure value and time penalty. The treasure value is 0 unless the agent reaches a treasure location. Each move returns -1 time penalty. The Pareto fronts including non-dominated solutions corresponding to two (3-column and 5-column) DST environments are depicted in Fig. 5.

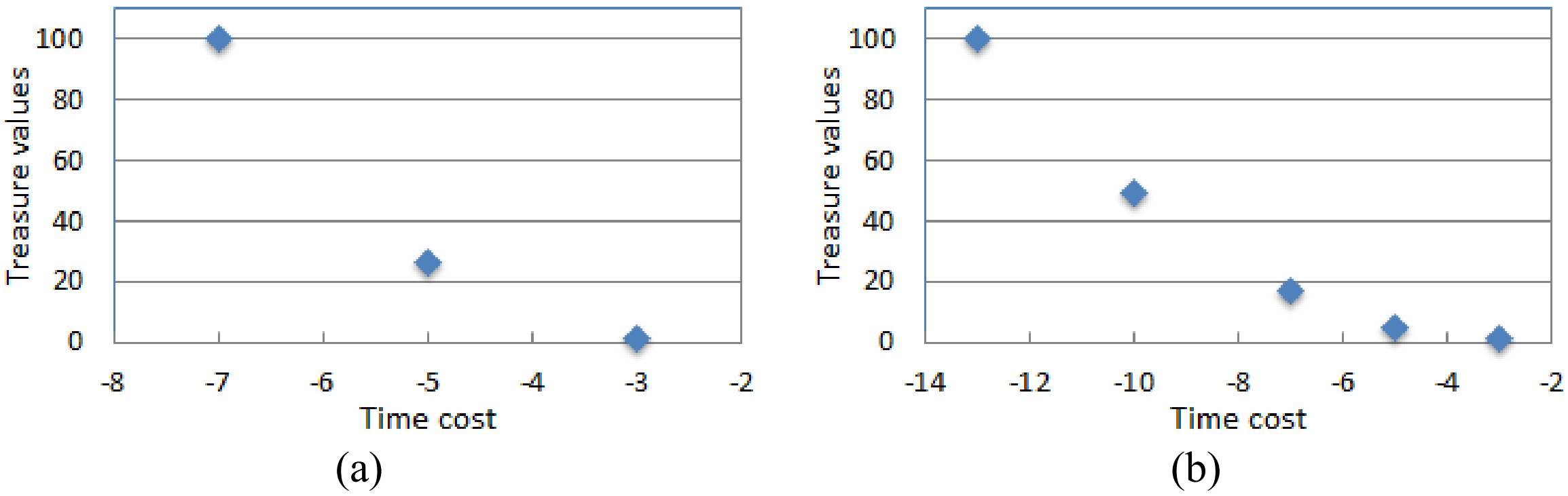

Fig. 5. The Pareto fronts for the DST problems, (a) 3 columns, and (b) 5 columns.

The following subsections present the experimental results of the 3-column environment, including both single-policy and multi-policy methods. The experimental results obtained using the MODRL framework on the 5-column DST environment are presented in Appendix A.

*4.1. Single-policy linear DQN*

Fig. 6 shows the convergence of the DQN-based MODRL framework applied to the 3-column DST environment. The solution (1, -3) is found after 100,000 steps using the linear scalarisation weights of [0.01, 0.99].

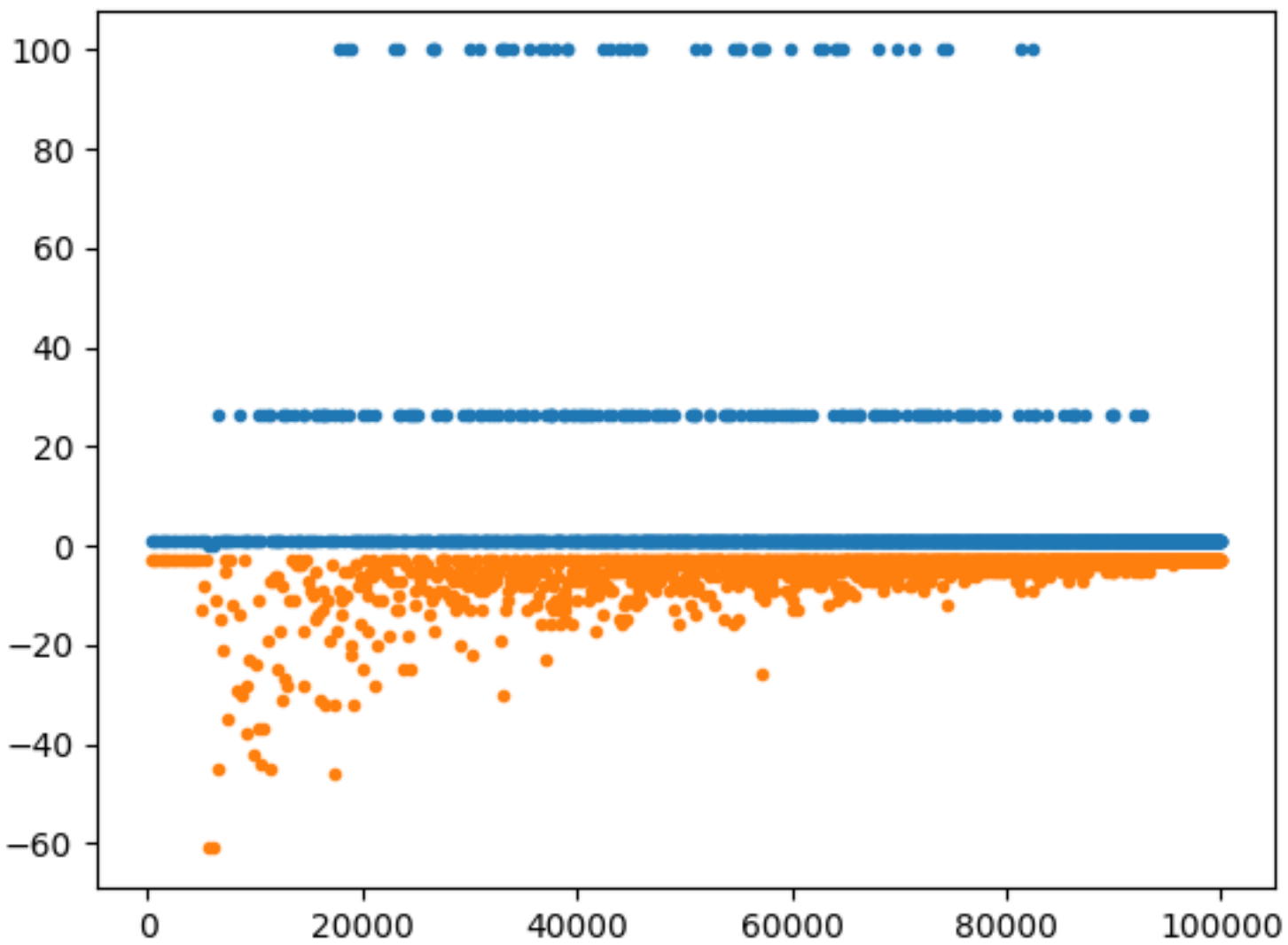

Fig. 6. Convergence of the learning process of the single policy linear DQN to solution (1, -3) of the 3-column DST environment. The $y$-axis represents the values of objectives (rewards) during the learning process, whilst the $x$-axis shows the number of training steps (actions) the agent has gone through. The blue dots show the rewards of the treasure objective whilst the yellow dots exhibit the rewards of the time penaly objective.

*4.2. Single-policy nonlinear DQN*

In this demonstration, we show that the linear approach cannot work with all cases. As an example, in the 3-column DST environment ($width = 3$), the possible Pareto solutions are (1, -3), (26.25, -5), and (100, -7). We can use the linear approach to direct the algorithm to find the solution (1, -3) using weights [0.01, 0.99] and solution (100, -7) using weights [0.5, 0.5], but it is impossible to find the second solution (26.25, -5) with any set of weights [a, b] where $a, b > 0$. Therefore, a non-linear approach, e.g. TLO, can be used. Fig. 7 illustrates the convergence to the optimal solutions achieved by the proposed framework using the nonlinear TLO method.

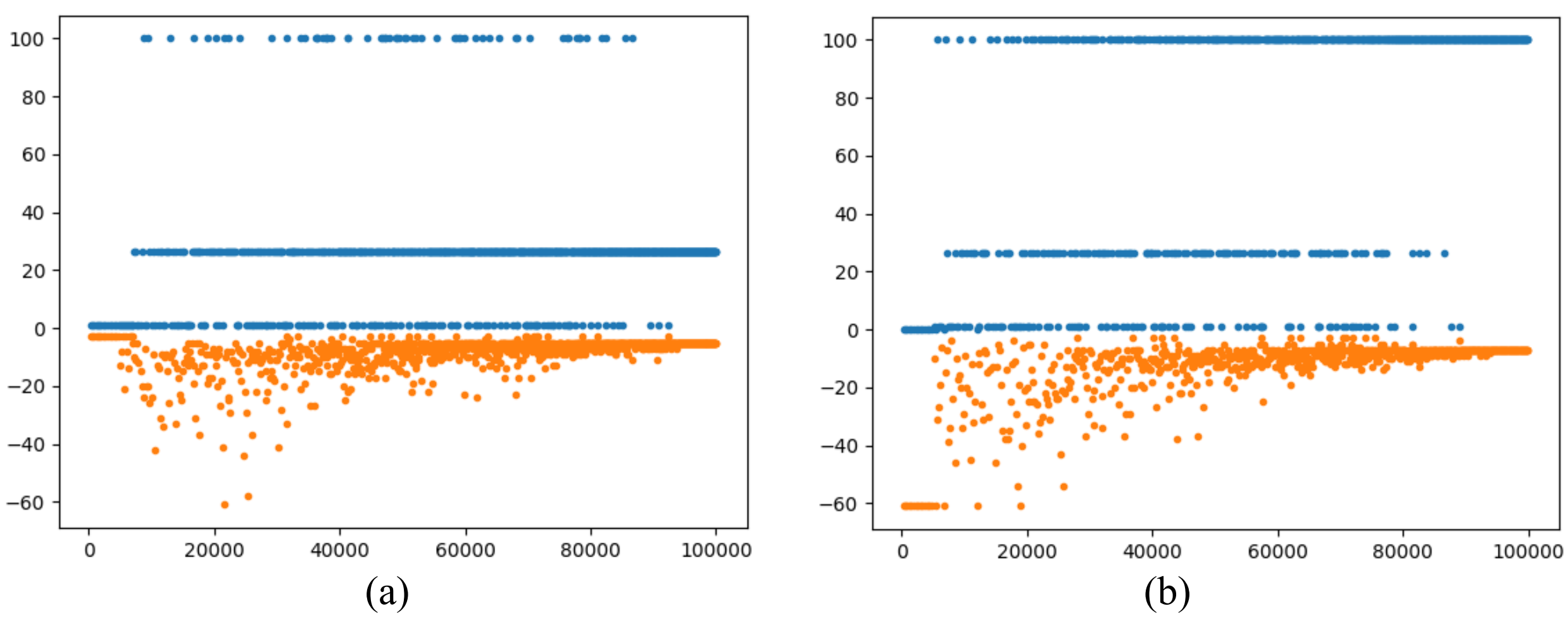

Fig. 7. (a) The learning process of nonlinear TLO DQN method converges to the solution (26.25, -5) of the 3-column environment with the threshold equal to 13.63, i.e. the average of 1 and 26.25. The threshold is only applied to the first objective (treasure). (b) Convergence of the TLO learning process to the solution (100, -7) with a threshold for the first objective equal to 63.13, i.e. the average of 26.25 and 100.

*4.3. Single-policy versus Multi-policy*

The framework is developed in such a way that multiple agents can be trained in parallel through multiple threads. It means that each agent is responsible for finding an individual optimal policy. Therefore, it is efficient to select a suitable policy when the required goal changes in real-world applications.

Fig. 8 shows the comparison between linear and nonlinear TLO agents using both single-policy and multi-policy methods. As discussed earlier, the nonlinear method can find all solutions in the concave front of the Pareto set while the linear method cannot converge to the solution (26.25, -5). Therefore, the nonlinear method performs better than the linear one. This result is aligned with the finding of Issabekov and Vamplew (2012). Using traditional MORL, Issabekov and Vamplew (2012) concluded that the nonlinear TLO outperforms the linear weighted sum method when the problem has no more than one intrinsic reward, e.g. the DST problem. From the obtained results, we can conclude that our implementation of deep learning extension preserves the properties of traditional MORL algorithms.

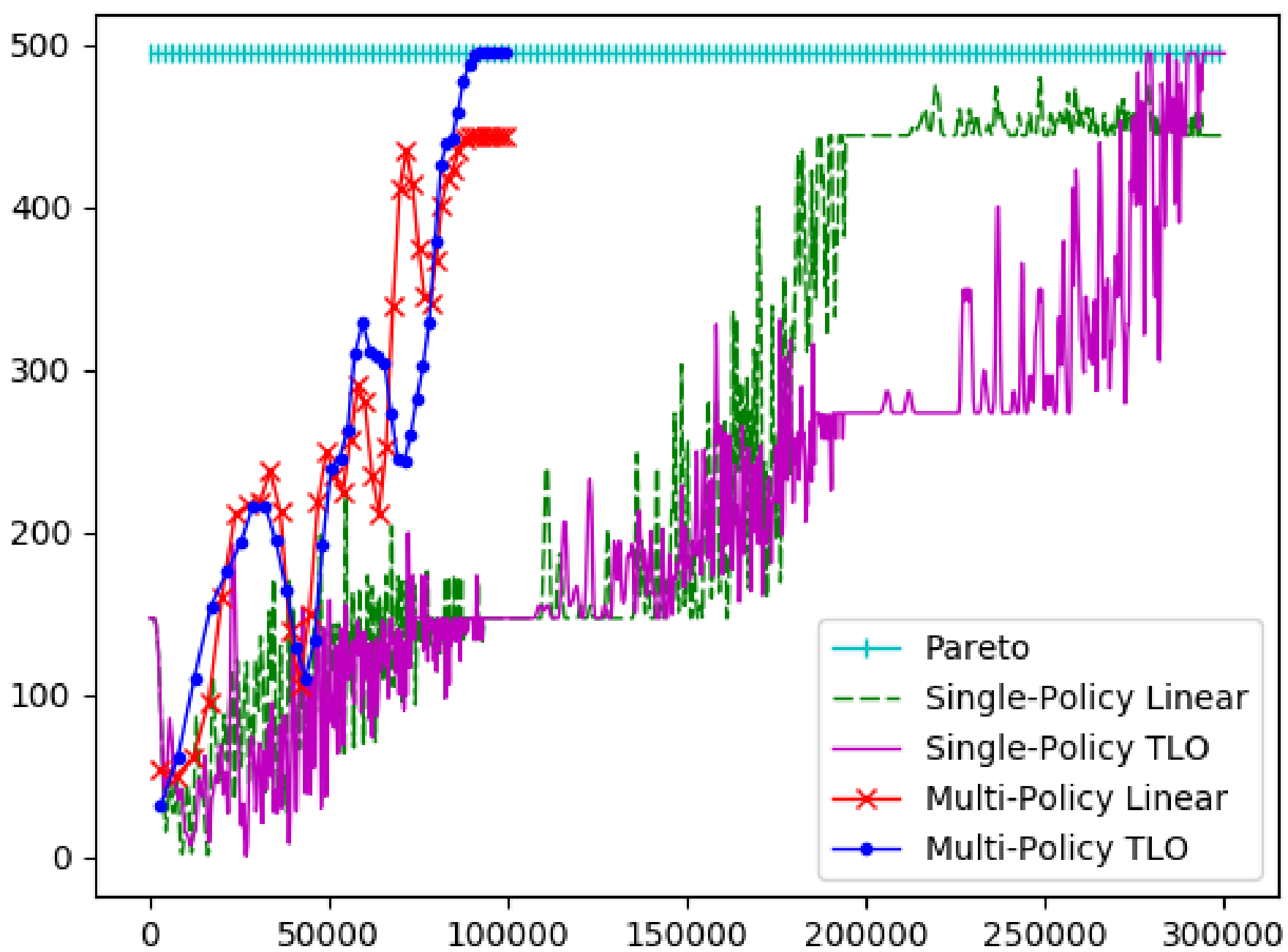

Fig. 8. History of the online hypervolume values of the approximation fronts learned by the linear and nonlinear TLO agents using both single-policy and multi-policy methods, and of the actual Pareto front (cyan). Approximately, at 300,000 steps (actions/moves), the single-policy nonlinear agent successfully found all three solutions, and therefore the hypervolume values of the approximation fronts (purple) are converged to that of the actual Pareto front. It is worth emphasising that the single-policy approach switches its weights after every 100,000 episodes. In contrast, the hypervolume values of the multi-policy nonlinear method (blue) converge to that of actual Pareto front at about 100,000 steps, which is approximately three times faster than that of the single-policy case where an agent sequentially finds all three solutions.

## 5. The MO Mountain Car Problem

The Mountain Car problem defines an environment where a car is required to escape from a valley as illustrated in Fig. 8. The car's engine is not powerful enough to climb up the mountain on the right side. Therefore, the car needs to reverse up the left side to obtain additional momentum. The learning algorithm's inputs are the car's current position and velocity whilst the action sets include forward acceleration, backward acceleration, and zero throttle (null action). The first objective of the problem is to minimize the number of steps taken by the car. Two other objectives include minimizing the number of backward and forward acceleration actions. As such, a penalty of -1 is applied to each time step and the same is applied to each backward (or forward) acceleration action. The MO Mountain Car problem has three intrinsic rewards corresponding to the three objectives. In the implementation, we limit the time steps to 100; therefore, an episode terminates when the time step exceeds 100 or the car reaches the goal.

In traditional Q-learning, there is a need to discretise the continuous state space to evaluate the optimal front (Alibekov et al., 2018). In the Mountain Car problem, the state space comprises the car’s position and velocity. With our implementation of the DQN, the state space is the entire image of the environment, as presented in Fig. 9. Note that the accrued values of the objectives (-100 -3 0) are just for demonstration, but they are not included in the image provided to the agent. This is the significant contribution of this study based on the integration of deep convolutional layers into the traditional Q-learning methods.

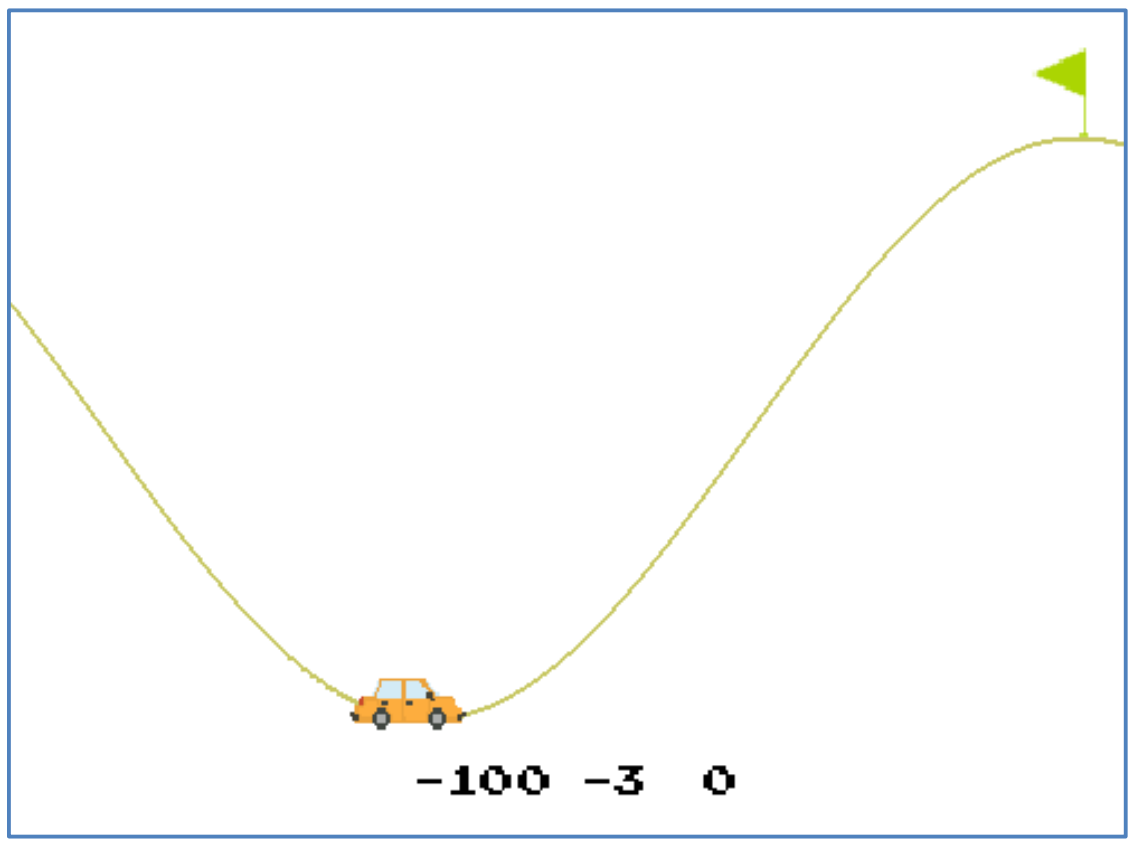

Fig. 9. Three-objective Mountain Car problem where the first objective of -100 represents time penalty, the second objective of -3 represents backward acceleration penalty, and the third objective of 0 represents the forward acceleration penalty.

*5.1. Single-policy linear DQN*

We evaluate the proposed framework on six different sets of weights, namely (1,0,0), (0.5,0.5,0), (0.5, 0, 0.5), (0, 1, 0), (0, 0.5, 0.5), and (0, 0, 1). The first element is the time step penalty, the second and third elements represent backward and forward acceleration penalties, respectively. The reward distributions obtained during training process of the linear weighted sum method with the 6 sets of weights are presented in Fig. 10.

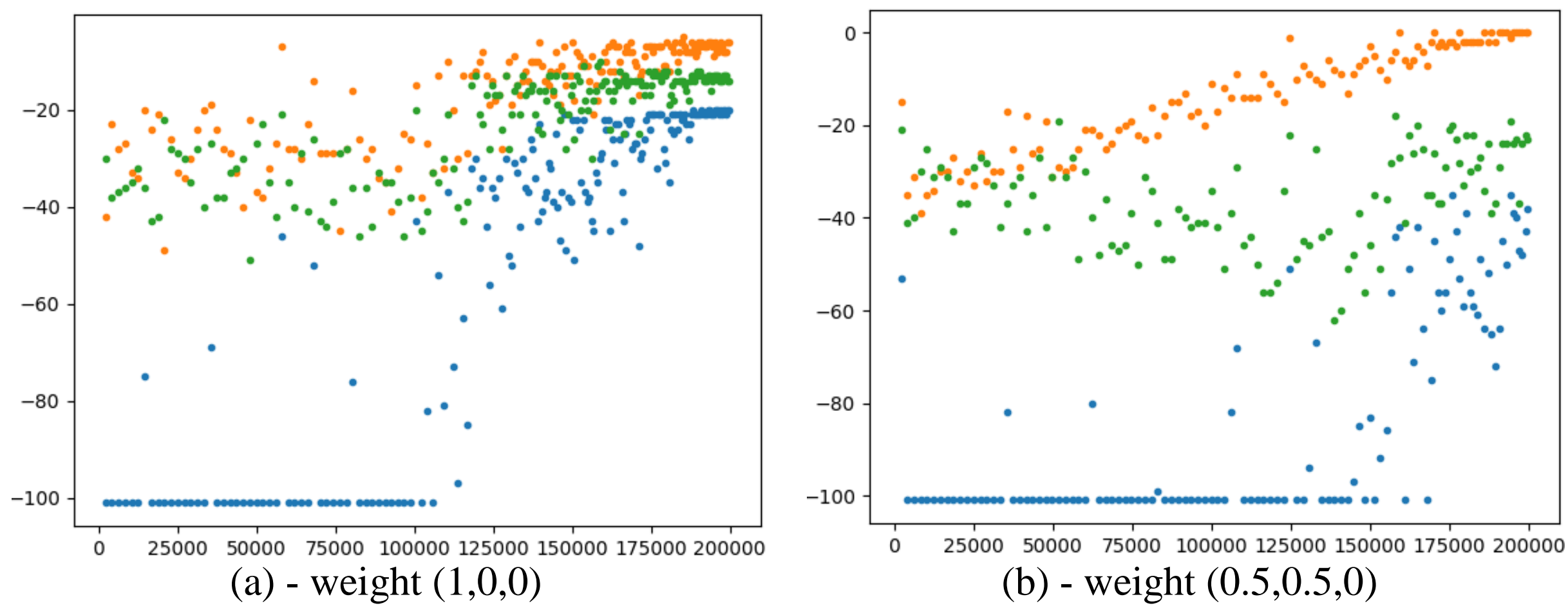

(a) - weight (1,0,0)  (b) - weight (0.5,0.5,0)

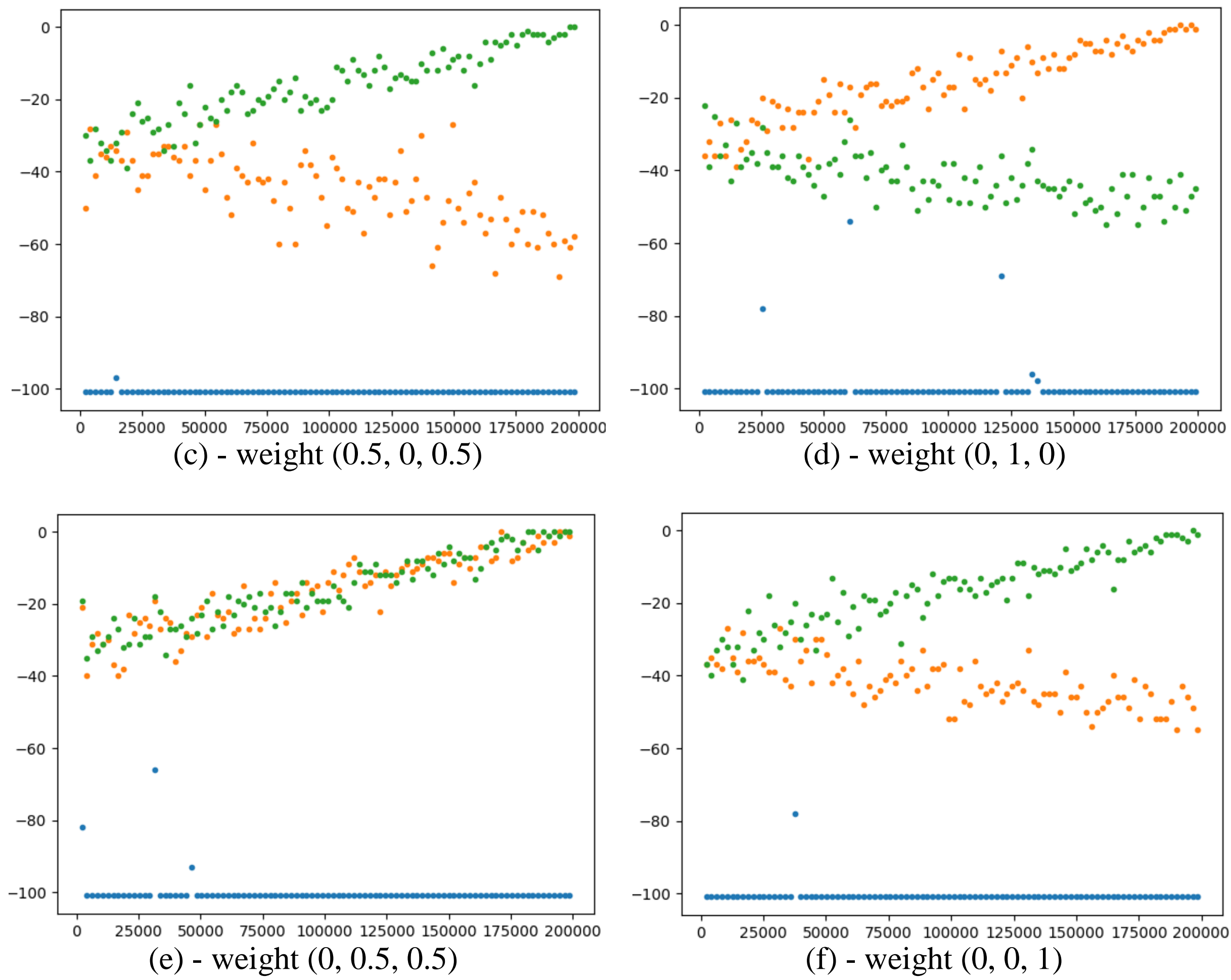

(c) - weight (0.5, 0, 0.5)

(d) - weight (0, 1, 0)

(e) - weight (0, 0.5, 0.5)

(f) - weight (0, 0, 1)

Fig. 10. The reward distribution of three objectives for the linear scalarisation DQN method with different sets of weights. The blue dots represent the time penalty whilst yellow and green dots are of the backward and forward acceleration penalties respectively.

When an objective is given a larger weight than those of the others, the algorithm tends to optimize that objective more effectively. In addition, we observe that if the weight for time is set too low, the agent may learn the policy in which it always chooses the null action, thereby incurring no penalty with regards to the other two objectives. This situation happens in the last three cases (0, 1, 0), (0, 0.5, 0.5), and (0, 0, 1). For example, in the case (0, 1, 0), the second objective (backward penalty – in yellow) is optimized to zero whilst two other objectives (blue and green) are not optimized. In contrast, in the case (0, 0, 1), the third objective (in green) is optimized to zero whilst the other two objectives (blue and yellow) are neglected. Noticeably, in the case (0, 0.5, 0.5), the second and third objectives (yellow and green) are both optimized effectively. This shows the adaptability and capability of the framework to converge to different optimal solutions.

*5.2. Single-policy nonlinear DQN*

With the nonlinear method for this three-objective problem, we evaluate our framework where the first two objectives (time and backward acceleration) are truncated with the following six TLO threshold cases: (0, -110), (-110, 0), (-110, -110), (-5, -3), (-5, -110), and (-110, -3). Fig. 11 shows the reward distributions of the three objectives after 200,000 training steps for these cases.

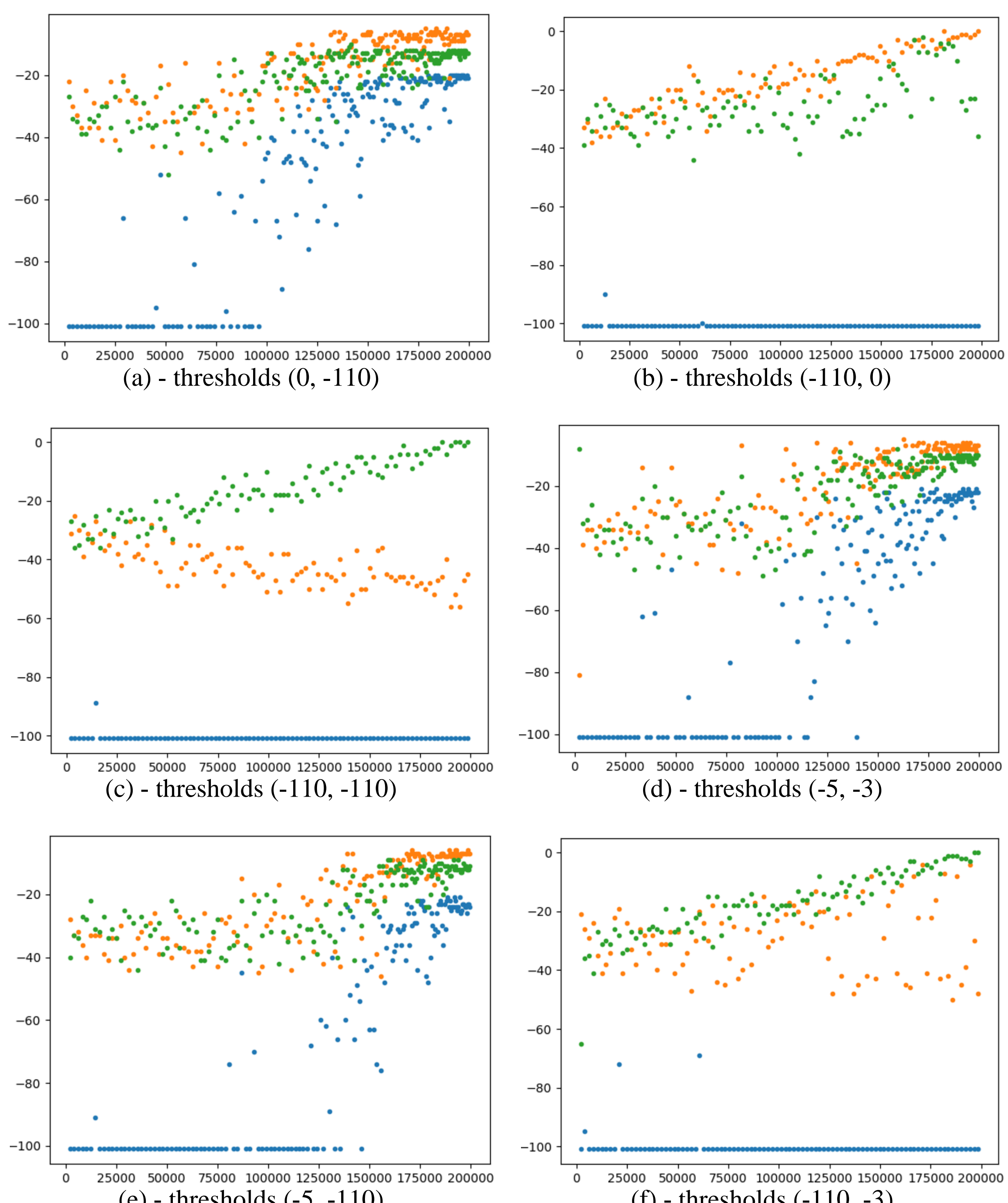

(a) - thresholds (0, -110)

(b) - thresholds (-110, 0)

(c) - thresholds (-110, -110)

(d) - thresholds (-5, -3)

(e) - thresholds (-5, -110)

(f) - thresholds (-110, -3)

Fig. 11. Reward distributions obtained when running the nonlinear TLO DQN method after 200,000 steps with different thresholds where blue, yellow and green are corresponding to time, backward and forward accelerations respectively.

Table 2. Summary of results of single-policy linear and non-linear DQNs

| **Linear Weights** | **Objectives Optimized** | **Rewards Obtained** | **Non-Linear Thresholds** | **Objectives Optimized** | **Rewards Obtained** |
|---|---|---|---|---|---|
| (1, 0, 0) | (1, 2) | (**-20**, **0**, -10) | (0, -110) | (1, 2) | (**-20**, **0**, -10) |
| (0.5, 0.5, 0) | (2) | (-40, **0**, -20) | (-110, 0) | (2) | (-100, **0**, -40) |
| (0.5, 0, 0.5) | (3) | (-100, -60, **0**) | (-110, -110) | (3) | (-100, -60, **0**) |
| (0, 1, 0) | (2) | (-100, **0**, -50) | (-5, -3) | (1, 2) | (**-20**, **0**, -10) |
| (0, 0.5, 0.5) | (2, 3) | (-100, **0**, **0**) | (-5, -110) | (1, 2) | (**-20**, **0**, -10) |
| (0, 0, 1) | (3) | (-100, -60, **0**) | (-110, -3) | (3) | (-100, -50, **0**) |

Table 2 summarizes results of single-policy linear and non-linear DQNs for sub-figures (a) to (f) of Figs. 10 and 11, as presented above. The column “Objectives Optimized” shows objectives that are optimized after learning using the corresponding weights (linear) or thresholds (non-linear). Specifically, ‘1’, ‘2’ and ‘3’ indicate time penalty, backward and forward acceleration penalties in the MO Mountain Car problem, respectively. The column “Rewards Obtained” represents the rewards that the agent obtained after learning. A reward in bold indicates the optimal value of individual objectives. For example, the reward (**-20**, **0**, -10) shows that objective 1 (time penalty) and objective 2 (backward acceleration penalty) are optimized and the agent after learning has obtained their optimal values of -20 and 0, respectively. We can see that different priority weights or thresholds lead to different objectives being optimized. The optimal values of objective 1 is -20 while those of objectives 2 and 3 are both equal to 0. The linear DQN method is more straightforward than non-linear version because in the linear version we only need to specify the priority weights (in range from 0 to 1) of objectives. In contrast, the non-linear version requires us to find out the ranges of objectives’ values before specifying appropriate thresholds.

*5.3. Single-policy versus Multi-policy*

Fig. 12 shows the history in terms of the hypervolume indicator of the multi-policy training process. With the multi-thread implementation, the algorithms (both linear and nonlinear TLO) require only 200,000 training steps to find 6 solutions simultaneously. The training steps required to run this experiment is 6 times smaller than that of the single-policy methods, i.e. 200,000 versus 1,200,000 training steps.

Both single-policy and multi-policy methods show the dominance of the linear weighted sum method against the nonlinear TLO method (Fig. 12). Our finding is again commensurate with that of Issabekov and Vamplew (2012), which shows that the TLO can only be effective for the problem with no more than one intrinsic objective, whereas for the Mountain Car problem, all three objectives are intrinsic. Under the implementation perspective, it is difficult to determine the thresholds because we need to observe the output range (min and max) of Q-values and then examine different sets of thresholds. In our MO Mountain Car experiments, we repeat the experiments with different threshold values and only select the thresholds that provide the best performance. Even with this experimental procedure, the nonlinear TLO method is still inferior to the linear method in solving the Mountain Car problem.

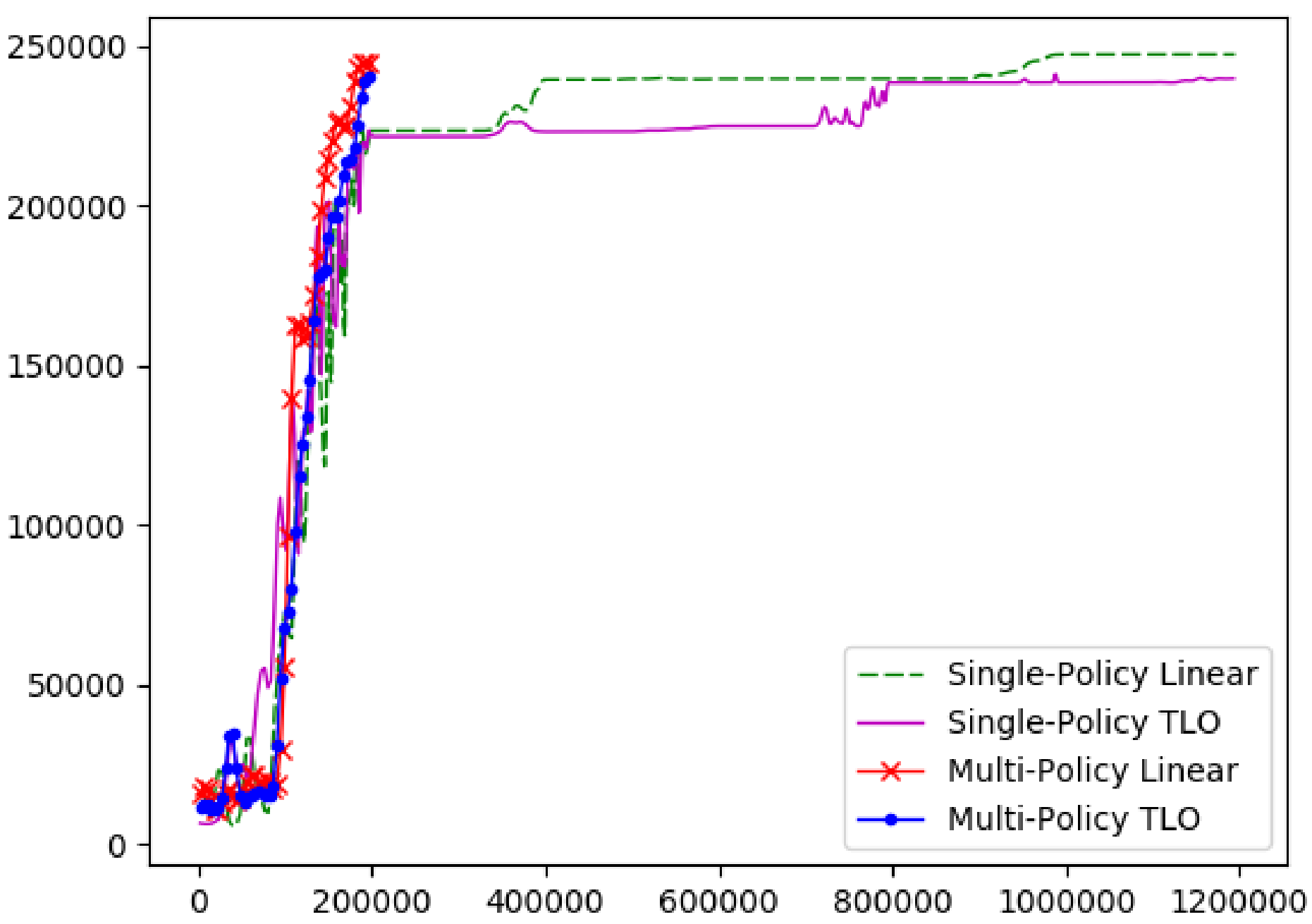

Fig. 12. Hypervolume values of linear and nonlinear methods using the single-policy and multi-policy DQN approaches. For the linear single-policy DQN method, we run through all 6 aforementioned sets of weights, one after another, and obtain the hypervolume values, as presented by the green dash line. For the nonlinear single-policy method, six threshold cases are carried out and their hypervolume values are shown by the purple line. The horizontal axis range between multi-policy and single-policy methods is different, indicating that the multi-threaded system learns faster.

## 6. Conclusions and Further Work

In this paper, a new scalable and high-performance MODRL framework has been proposed. Its implementation using Python has been demonstrated. The integration of DRL algorithms into traditional MORL methods is important because such traditional methods, like tabular Q-learning, are not able to deal with high-dimensional environments. The proposed MODRL framework facilitates the use of both single-policy and multi-policy strategies to solving MORL problems efficiently. Our proposed framework utilizes multithreading to significantly reduce training time when dealing with multi-policy tasks. Most importantly, the framework is generic, highly modularized, and is able to accommodate different DRL algorithms, e.g. DQN, Dueling DQN, asynchronous advantage actor-critic (A3C), Double DQN (Nguyen et al., 2017), and UNsupervised Reinforcement and Auxiliary Learning (UNREAL) (Jaderberg et al., 2016), in various environments, from simple problem domains such as gridworlds, deep sea treasure, and Mountain Car to complicated problems, including Atari series and MuJoCo (Todorov et al., 2012; Duan et al., 2016). This constitutes one of our future research directions in expanding and enhancing the proposed MODRL framework. We also intend to investigate deep implementations of more multi-policy approaches which enable further efficiencies by updating multiple policies on each timestep using off-policy learning. Another further work will focus on developing multi-agent environments that can be integrated into the current framework to solve various problems of multi-agent-based systems.

There are many problems in the real world, especially in engineering, where the MODRL methods can be applied to. For example, an important application of MODRL is to control autonomous vehicles where objectives such as performance of vehicles and energy consumption are conflicting. We need to maximize the performance of vehicles while minimizing the energy consumption. In chemical engineering, a typical industrial setting can aim to minimize the cost of production while satisfying specifications on the products and effluents. In controlling an electric power distribution system, the objectives are to minimize the operating costs, pollution and atmospheric emissions while meeting the demand and satisfying the system constraints. Therefore, the proposal of the MODRL framework in this paper would contribute to solving various real-world engineering problems effectively.

**Appendix A**

The results of the MODRL framework on the 5-column DST problem are presented. This environment requires the agent to find 5 optimal policies of the actual Pareto front, including (1, -3), (5, -5), (17, -7), (49, -10), and (100, -13). Figures A1 to A4 illustrate the convergence of the learning process to these optimal solutions. The first solution, i.e. (1, -3), is found by the linear scalarisation method, while the remaining solutions are found by the nonlinear TLO method.

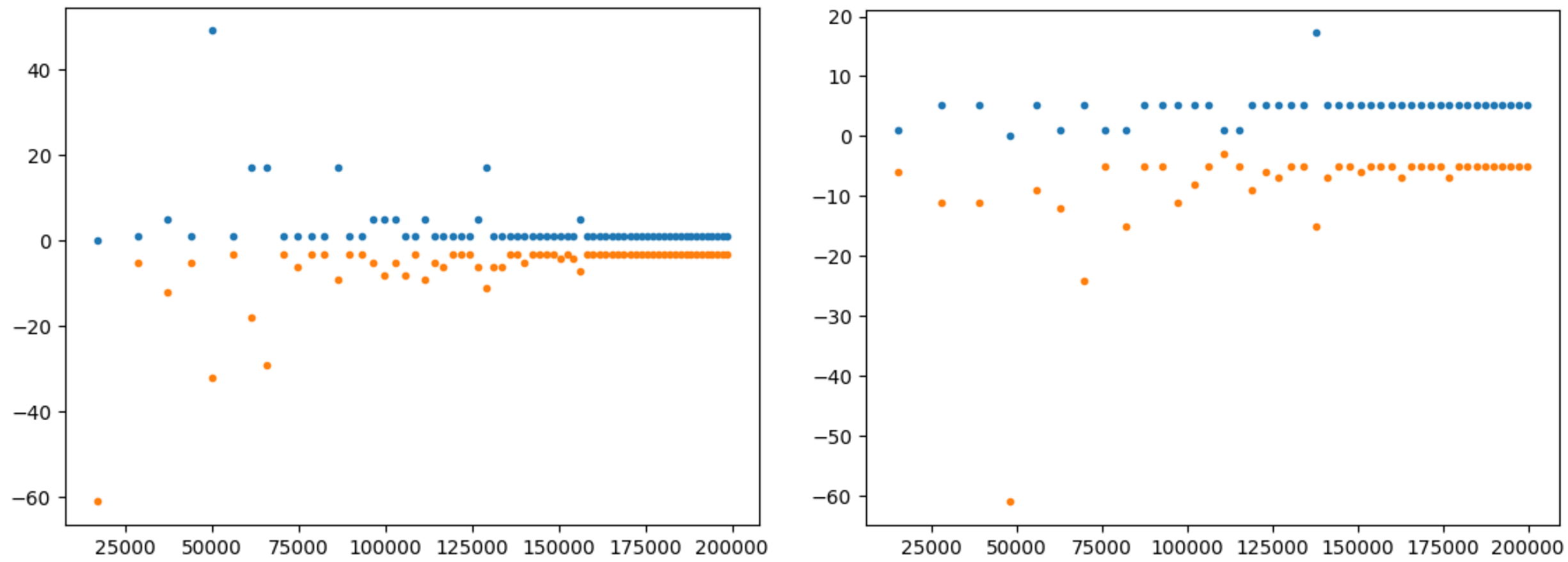

Fig. A1. Convergence of the MODRL learning process to the solutions (1, -3) (left), and (5, -5) (right).

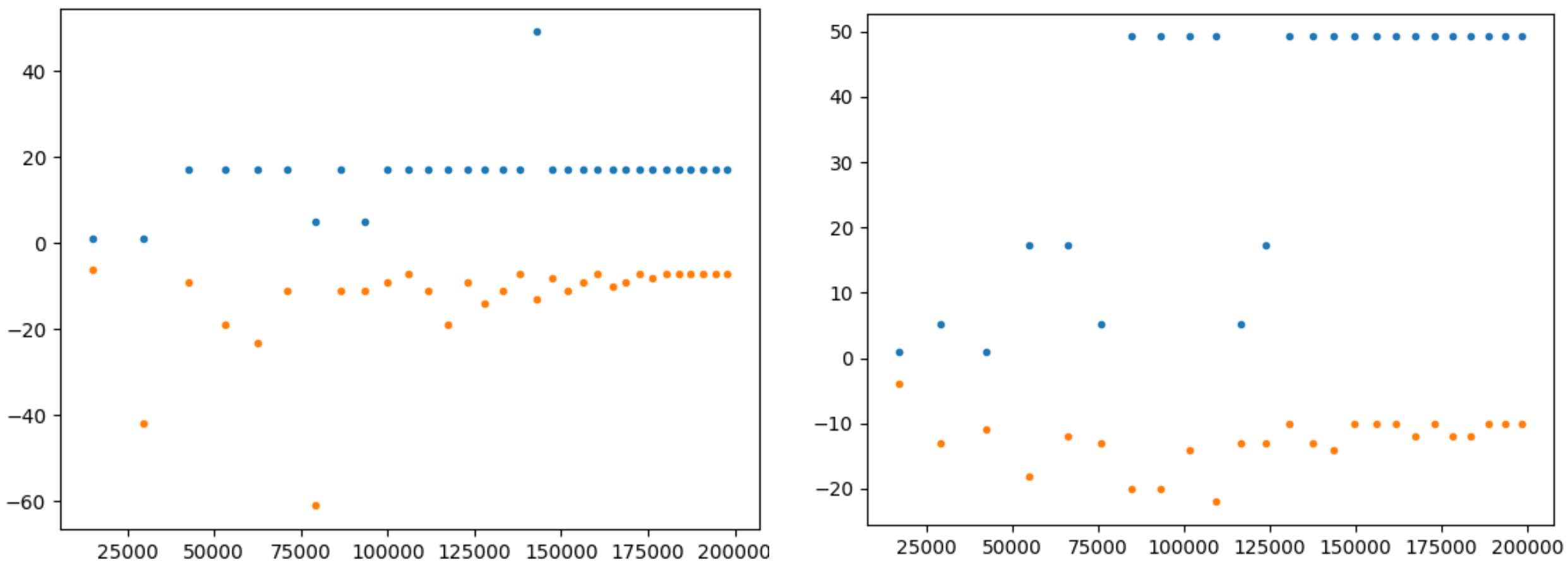

Fig. A2. Convergence of the learning process to solutions (17, -7) (left), and (49, -10) (right).

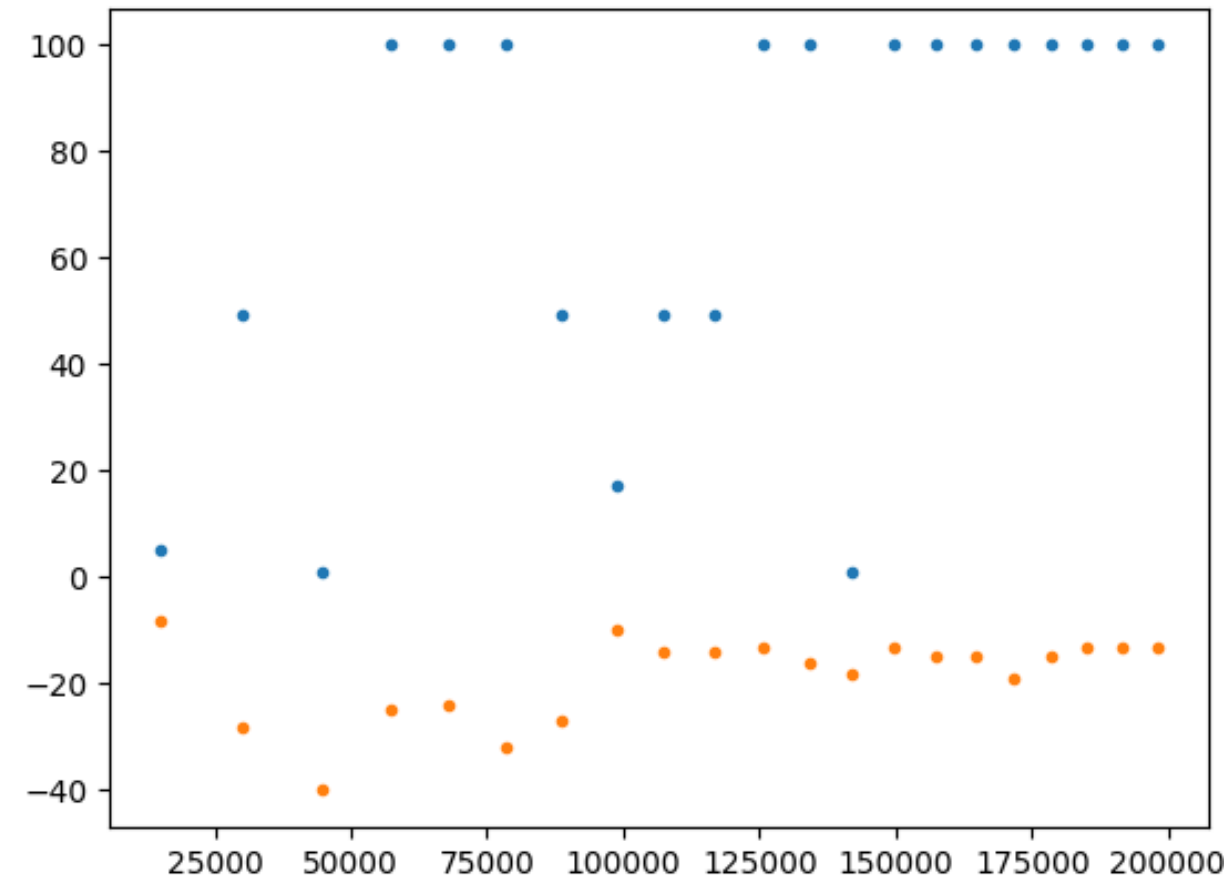

Fig. A3. Convergence of the learning process to the solution (100, -13).

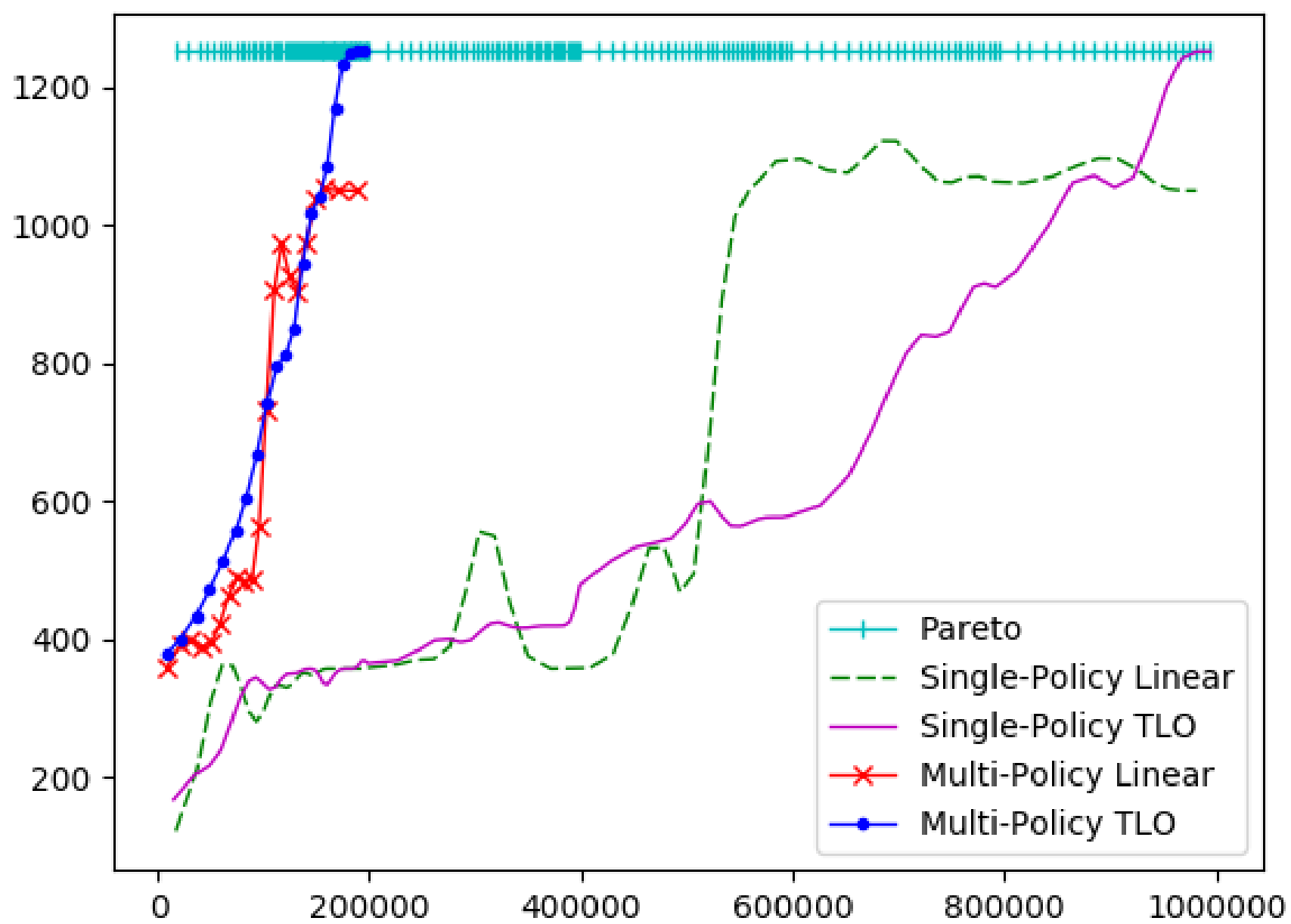

Fig. A4. Online hypervolume values obtained during the learning process of single-policy and multi-policy methods using both linear method and nonlinear TLO algorithm.